\documentclass[runningheads]{llncs}
\usepackage[T1]{fontenc}
\usepackage{color}
\usepackage{graphicx}
\usepackage{hyperref}
\usepackage[misc]{ifsym}
\usepackage{latexsym}
\usepackage{mathtools}
\usepackage{multirow}
\usepackage{subfigure}

\DeclareMathOperator*{\argmax}{arg\,max}
\begin{document}
\title{Enhancing Document-level Relation Extraction by Entity Knowledge Injection\thanks{National Natural Science Foundation of China (No. 61872172).}}
\titlerunning{Enhancing Document-level Relation Extraction}
%

\author {
    Xinyi Wang\inst{1} \and
    Zitao Wang\inst{1} \and
    Weijian Sun\inst{3} \and
    Wei Hu\inst{1,2}\textsuperscript{(\Letter)}
}
\authorrunning{X. Wang et al.}
%
\institute {
        State Key Laboratory for Novel Software Technology,\\ Nanjing University, Nanjing, China \\
\and    National Institute of Healthcare Data Science,\\ Nanjing University, Nanjing, China \\
\and    Huawei Technologies Co., Ltd., Shanghai, China \\
\email{\{xywang.nju,ztwang.nju\}@gmail.com, sunweijian@huawei.com, whu@nju.edu.cn}
}
\maketitle              
\begin{abstract}
Document-level relation extraction (RE) aims to identify the relations between entities throughout an entire document. It needs complex reasoning skills to synthesize various knowledge such as coreferences and commonsense. Large-scale knowledge graphs (KGs) contain a wealth of real-world facts, and can provide valuable knowledge to document-level RE. In this paper, we propose an entity knowledge injection framework to enhance current document-level RE models. Specifically, we introduce coreference distillation to inject coreference knowledge, endowing an RE model with the more general capability of coreference reasoning. We also employ representation reconciliation to inject factual knowledge and aggregate KG representations and document representations into a unified space. The experiments on two benchmark datasets validate the generalization of our entity knowledge injection framework and the consistent improvement to several document-level RE models.

\end{abstract}

\section{Introduction}
\label{sect:intro}
Relation extraction (RE) aims to recognize the semantic relations between entities in texts, which is beneficial to a variety of AI applications such as language understanding and knowledge graph (KG) construction.
Early methods \cite{heist2017language,zeng2014relation,zhang2018graph} mainly cope with sentence-level RE, which detects the relations in a single sentence. 
However, a large number of relations span across multiple sentences \cite{yao2019docred}, which calls for document-level RE in recent years.
Compared with sentence-level RE, document-level RE is more challenging.
It needs the RE models to conduct complex reasoning, e.g., coreference reasoning, factual reasoning and logical reasoning, throughout an entire document.

\begin{figure}[!tb]
	\centering
	\includegraphics[width=.9\columnwidth]{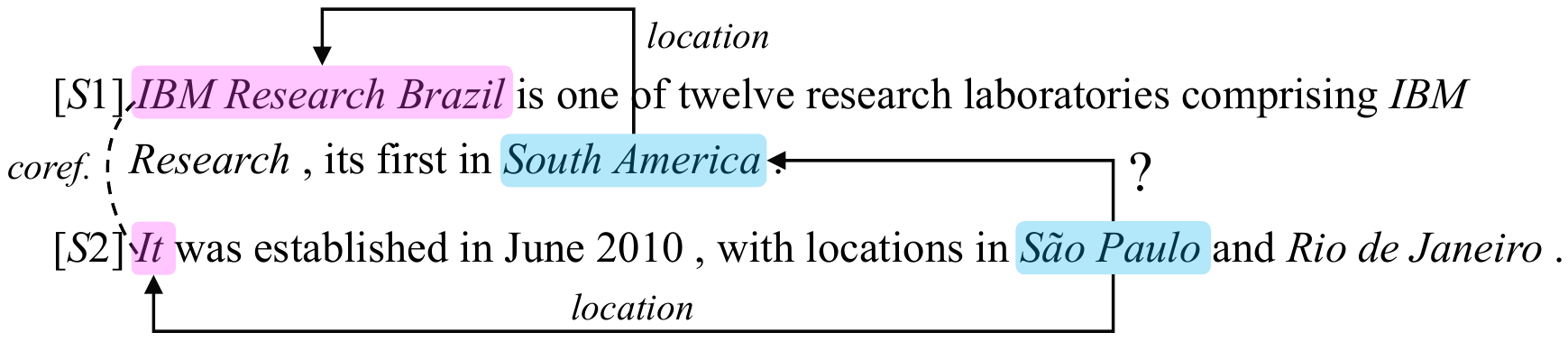}
	\caption{An example of document-level RE excerpted from \cite{yao2019docred}}
	\label{fig:example}
\end{figure}

Figure~\ref{fig:example} shows a real example. 
A document-level RE model is asked to find the relations between three named entities \textit{IBM Research Brazil}, \textit{S\~{a}o Paulo} and \textit{South America}. 
From $S1$, \textit{IBM Research Brazil} is located in \textit{South America} may be first recognized by the model. 
Then, with the help of coreference knowledge that connects the pronoun \textit{It} in $S2$ to \textit{IBM Research Brazil} in $S1$, the model can recognize that \textit{IBM Research Brazil} is located in \textit{S\~{a}o Paulo}. 
Since the model may not know the exact types of entities, only with the aid of extra knowledge in KGs like \textit{S\~{a}o Paulo} is a city and \textit{South America} is a continent, then it can confidently determine that the relation between them is \textit{continent} rather than others. 
The entire reasoning process demands the document-level RE model to synthesize various knowledge and have powerful reasoning capabilities. 

Recent years have witnessed that large-scale KGs, e.g., Wikidata
\cite{vrandevcic2014wikidata} and DBpedia \cite{auer2007dbpedia}, become a valuable asset in information extraction \cite{bastos2021recon,fernandez2020enhancing,pan2019entity,tong2020improving,turker2020weakly,vashishth2018reside,verlinden2021injecting}. 
A KG contains a collection of real-world facts, in which a fact is structured in the form of a triple $(\textit{entity}, \textit{property}, \textit{value})$. 
\textit{Property} can be either an \textit{attribute} or a \textit{relation}, and \textit{value} can be either a \textit{literal} for attribute triple or an \textit{entity} for relation triple.
Particularly for the RE task, the works in \cite{bastos2021recon,ji2017distant,vashishth2018reside,verlinden2021injecting} exploit one or very few attribute and relation triples (e.g., \textit{rdfs:label}) in KGs to enhance their models.
Furthermore, they overlook the heterogeneity between KG representations and document representations, and aggregate them in a simple way like vector concatenation.

In this paper, we propose a novel entity knowledge injection framework to enhance existing document-level RE models.
Specifically, we introduce a general \emph{knowledge injection layer} between the encoding layer and the prediction layer of popular RE models.  
Based on it, we focus on injecting various entity knowledge from KGs into the document-level RE models. 
We tackle two key challenges:

First, \emph{how to inject coreference knowledge into document-level RE models?}
Coreference resolution plays a vital role in RE. 
However, the coreferences derived from coreference resolution tools and aliases in KGs may contain errors. 
If we directly import them into an RE model as strong guidance information, such as the edges in a document graph \cite{quirk2017distant}, it is likely to bring a downside effect. 
Therefore, we present \emph{coreference distillation} to distill knowledge from the coreferences and inject it into an RE model, so that the model can ultimately acquire generalized coreference knowledge.

Second, \emph{how to inject factual knowledge into document-level RE models?}
KG contains a wealth of facts related to entities, which we want to exploit for RE. 
However, the representations of entities in a KG and the text representations of a document are learned in two different spaces, which demand to be reconciled together. 
We present \emph{representation reconciliation} to fuse KG representations and document representations into a unified space, endowing the RE model with the factual knowledge of entities.

In summary, our main contributions in this paper are twofold:
\begin{itemize}
\item We define a general knowledge injection framework KIRE and design various knowledge injection tasks for document-level RE, such as coreference distillation for coreference knowledge and representation reconciliation for factual knowledge.
These knowledge injection and RE tasks are optimized together by multi-task learning. (Sections~\ref{sect:problem} and \ref{sect:model})

\item We perform the experiments on two benchmark datasets DocRED \cite{yao2019docred} and DWIE \cite{zaporojets2021dwie} for document-level RE.  
The result comparison between seven RE models and the models after knowledge injection validates the generalization and stable improvement of our framework. (Section~\ref{sect:exp})
\end{itemize}


\section{Related Work}
\label{sect:work}

\textbf{Document-level RE.} Document-level RE has attracted vast attention in the past few years. 
A considerable number of studies have been conducted, which can be generally divided into graph-based models \cite{quirk2017distant,peng2017cross,verga2018simultaneously,nan2020reasoning,wang2020global,zeng2020double,xu2021document} as well as sequence-based models \cite{wang2019fine,tang2020hin,huang2021three,zhou2021document,xu2021entity}.
Graph-based models build document graphs to capture the semantic information in a document, and design various neural networks to carry out inference on the built document graphs.
DISCREX \cite{quirk2017distant} models words in a document as nodes and intra/inter-sentential dependencies as edges.
Following this idea, Peng et al. \cite{peng2017cross} make use of graph LSTM while BRAN \cite{verga2018simultaneously} employs Transformer to encode document graphs.
Recently, LSR \cite{nan2020reasoning}, GAIN \cite{zeng2020double} and GLRE \cite{wang2020global} define more sophisticated document graphs to reserve more dependency information in a document.

Sequence-based models adopt neural encoders like BERT to implicitly capture dependencies in a document, instead of explicitly building document graphs.
Wang et al. \cite{wang2019fine} use BERT to encode a document and design a two-step pipeline, which predicts whether a relation exists between two entities first, and then predicts the specific relation types.
HIN \cite{tang2020hin} also makes use of BERT but design a hierarchical model that integrates the inference information from the entity, sentence and document levels.
Huang et al. \cite{huang2021three} extract three types of paths which indicate how the head and tail entities can be possibly related in the context, and predict the relations based on the extracted evidence sentences.
ATLOP \cite{zhou2021document} proposes localized context pooling to transfer attentions from pre-trained language models and adaptive thresholding to resolve the multi-label and multi-entity problem.
SSAN \cite{xu2021entity} modifies the attention mechanism in BERT to model the coreference and co-occurrence structures between entities, to better capture the semantic information in the context.

In this paper, our focus is injecting knowledge into these document-level RE models. 
Our entity knowledge injection framework KIRE is applicable to various models as long as they fall into our framework formulation.

\textbf{Knowledge injection.} A few works have studied how to inject external knowledge such as a KG into the RE task for performance improvement.
RESIDE \cite{vashishth2018reside} uses entity types and aliases while $\text{BERT}_\text{EM+TM}$ \cite{fernandez2020enhancing} only uses entity types.
They both consider very limited features of entities.
RECON \cite{bastos2021recon} proposes separate models to encode attribute triples and relation triples in a KG and obtain corresponding attribute context embeddings and relation context embeddings, which are combined into sentence embeddings. 
KB-both \cite{verlinden2021injecting} utilizes entity representations learned from either hyperlinked text documents (Wikipedia) or a KG (Wikidata) to raise the information extraction performance including document-level RE. 
Different from all above, we integrate more types of knowledge including coreferences, attributes and relations symbiotically with more effective knowledge injection methods to address the document-level RE task.

Additionally, a few studies \cite{liu2020kbert,wei2021knowledge,zhang2019ernie} explicitly exploit incorporating knowledge from various sources such as encyclopedia knowledge, commonsense knowledge and linguistic
knowledge into pre-trained language models with different injection strategies to improve the performance of language models in downstream tasks.
However, the goal of these studies is orthogonal to this paper.

\section{Framework Formulation}
\label{sect:problem}

According to \cite{tang2020hin,yao2019docred,zhou2021document}, we formulate the document-level RE task as a \emph{multiple binary classification} problem.
Given a document annotated with entities and their corresponding textual mentions, the task aims to predict the relations for each entity pair in the document, where a relation is either a predefined type (e.g., \textit{country}) or \textit{N/A} for no relation.
Note that there may be more than one relation for an entity pair.

A basic neural network model \cite{yao2019docred} for document-level RE contains an encoding layer and a prediction layer.
The encoding layer encodes an input document to obtain the context-sensitive representations of tokens (words) in it, and the prediction layer generates entity representations and predicts relations using the entity representations.
In this paper, we add a \emph{knowledge injection layer} between the encoding layer and the prediction layer, and many document-level RE models such as \cite{wang2020global,yao2019docred,zhou2021document} can be used as the basic model. 

We regard a KG as the knowledge source for injection.
A KG is defined as a 7-tuple $\mathcal{G}=(U,R,A,$ $V,X,Y,C)$, where $U,R,A$ and $V$ denote the sets of entities, relations, attributes and literal values, respectively. $X \subseteq U \times R \times U$ denotes the set of relation triples, $Y \subseteq U \times A \times V$ denotes the set of attribute triples, and $C$ denotes the set of coreference triples derived from $\mathcal{G}$.
By the alias information (e.g., \textit{skos:altLabel}) in $\mathcal{G}$, any two aliases of an entity can constitute one coreference triple $(m_s,m_t,p_{cr})$, where $m_s,m_t$ are two alias mentions and $p_{cr}$ is the coreference probability.
We employ off-the-shelf coreference resolution models to find more coreference knowledge for pronouns (e.g., \textit{it} and \textit{he}), possessives (e.g., \textit{herself}), noun phrases (e.g., \textit{this work}), etc., in the document.
$p_{cr}$ is set to the resolution confidence.
Due to the main scope of this paper, we follow \cite{liu2020kbert,zhang2019ernie} and reuse entity linking tools to link the entities in the document to those in the KG. 

\textbf{Framework.}
Given a document $\mathcal{D}=\{w_1,\dots,w_J\}$, where $w_j$ denotes the $j^\text{th}$ token in $\mathcal{D}$, and a KG $\mathcal{G}$, the framework of document-level RE with entity knowledge injection is
\begin{align}
	\begin{aligned}
		\mathbf{H} &= [\mathbf{h}_{w_1},\dots,\mathbf{h}_{w_J}] = \mathrm{Encode}(\mathcal{D}), \\
		\mathbf{H}' &= \mathrm{KnowledgeInject}(\mathcal{D},\mathbf{H},\mathcal{G}), \\
		\mathbf{z} &= \mathrm{Predict}(\mathbf{H}'),
	\end{aligned}
\end{align}
where $\mathbf{h}_{w_j}$ denotes the hidden representation of $w_j$, and $\mathbf{z}$ denotes the prediction probability distribution of relations. \hfill $\Box$

\section{Knowledge Injection}
\label{sect:model}

\begin{figure}[!tb]
	\centering
	\includegraphics[width=\columnwidth]{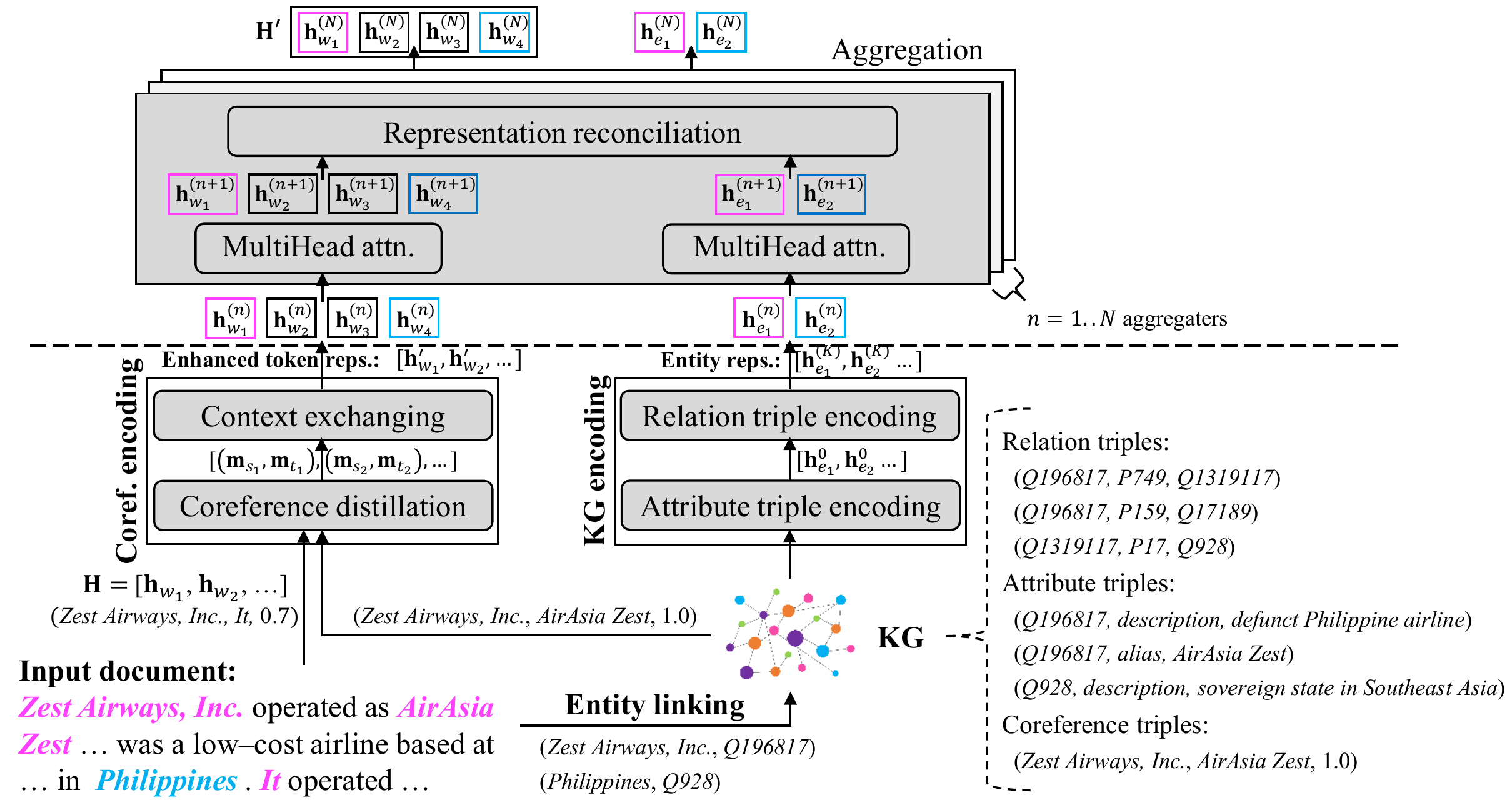}
	\caption{Architecture of the knowledge injection layer}
	\label{fig:model}
\end{figure}

The architecture of the proposed knowledge injection framework KIRE is depicted in Figure~\ref{fig:model}, which accepts the document $\mathcal{D}$, the hidden representation $\mathbf{H}$ of $\mathcal{D}$ and the relevant KG $\mathcal{G}$ as input. 
It injects the entity knowledge from the coreference triples, attribute triples and relation triples into an RE model, and outputs the final hidden representation $\mathbf{H}'$. 

Specifically, we inject the coreference triples into the basic document-level RE model with coreference distillation and context exchanging. 
Apart from this, the attribute triples are semantically encoded with AutoEncoder \cite{rumelhart1986learning}, and the encoded results are then reused to initialize the representations of relation triples.
We use a relational graph attention network (R-GAT) \cite{busbrige2019relational} to encode the relation triples and generate the KG representations of entities.
Finally, the KG representations of entities and the token representations that have been enhanced by coreference knowledge are aggregated by representation reconciliation. 
Details are described in the following subsections.

\subsection{Coreference Encoding}
This module leverages coreference triples to exchange the contextual information between aliases, and thus the representations of alias mentions can be closer. 

\textbf{Coreference Distillation.}
A simple method is to model the coreference triples as a new type of edges in the document graph and reuse graph-based models \cite{nan2020reasoning,peng2017cross,verga2018simultaneously}.
However, such a method cannot be generalized to the sequence-based models since they do not construct document graphs.
Furthermore, the accuracy of existing coreference resolution tools is still far from perfect, even they are trained on large corpora. 
To alleviate error accumulation, it is inappropriate to directly add the edges as strong guidance information in the RE models.

Knowledge distillation \cite{hinton2015distilling,tong2020improving}, as a model compression technique and a solution to integrate external knowledge into a target model, has been used in a wide range of NLP tasks. 
In this paper, we leverage the idea of knowledge distillation and propose coreference distillation to inject coreference triples into the RE models.
Our main idea is to leverage a pre-trained coreference resolution model which has been trained on a large coreference dataset as the \emph{teacher} model, and then force the \emph{student} model (i.e., the RE model) to generate a prediction probability distribution that approximates the teacher model on the coreference triples.
Finally, the student model learns the coreference knowledge and generalization ability in the teacher model.
Formally, for a coreference triple $(m_s,m_t,$ $p_{cr})$, its coreference probability generated by the teacher model is defined as
\begin{align}
	P_\text{tea}(m_s,m_t) = p_{cr}.
\end{align}

The student model generates the coreference probability with a multi-layer perceptron (MLP):
\begin{align}
	P_\text{stu}(m_s,m_t) = \mathrm{MLP}\Big(\big[\mathbf{m}_s;\mathbf{m}_t; \mathbf{\Delta} (\psi (m_s,m_t))\big]\Big),
\end{align}
where $\mathbf{m}_s$ and $\mathbf{m}_t$ denote the hidden representations of alias mentions $m_s$ and $m_t$, respectively, which are calculated by averaging the hidden representations of tokens in $m_s$ and $m_t$,
that is, $\mathbf{m}_s = \mathrm{avg}_{w_j \in m_s}(\mathbf{h}_{w_j}), \mathbf{m}_t = \mathrm{avg}_{w_j \in m_t}(\mathbf{h}_{w_j})$.
``;'' is the concatenation operation,
and $\psi (m_s,m_t)$ denotes the shortest distance between $m_s,m_t$ in the document.
We divide the distance into $\{1,2,\dots,2^\beta\}$ bins, and associate each bin with a trainable distance vector.
$\mathbf{\Delta} (\cdot)$ associates each $\psi$ to the distance vector of relevant bin. Empirically, aliases with different distances should have different impacts on each other. Therefore, we propose trainable distance vectors to model and utilize such difference.

We enforce the student model to learn from the teacher model using the following coreference loss:
\begin{align}
    \mathcal{L}_{cr} = \sum_{(m_s,m_t)\in C} \mathrm{KL} \big(P_\text{tea}(m_s,m_t) \parallel P_\text{stu} (m_s,m_t)\big),
\end{align}
where $\mathrm{KL}(\cdot)$ is the Kullback-Leibler divergence.

\textbf{Context exchanging.}
Based on the learned coreference knowledge, we further enable each alias mention to interact with its most similar counterpart, so as to exchange the semantic information between them.
Specifically, given an alias mention $m_s$, we update its hidden representation through $\mathbf{m}_s = \mathbf{m}_s + \mathbf{m}_{t^*}$,
where $t^* =$ $\argmax_t \big\{P_\text{stu} (m_s, m_t) \,|\,(m_s,m_t) \in C\big\}$. 
In this way, the representations of the pronouns in particular can be enriched via their referents.

Finally, we obtain the token representations enhanced by coreference knowledge through the representations of alias mentions (if exists): 
\begin{align}
    \mathbf{h}'_{w_j} = \left\{
	\begin{array}{ll}
		\mathbf{m}_s, 		& \text{if } w_j \in m_s \\
		\mathbf{h}_{w_j},	& \text{otherwise} \\
	\end{array}.
\right.
\end{align}

In coreference encoding, the MLP contains $d_\text{MLP}(2d_\text{token}+d_\text{dist})+2d_\text{token}$ parameters, where $d_\text{MLP},d_\text{token},d_\text{dist}$ are the dimensions of MLP hidden layers, token representations and trainable distance vectors, respectively.

\subsection{Knowledge Graph Encoding}
This module aims to encode the attribute triples and the relation triples to generate the KG representations of entities.

\textbf{Attribute triple encoding.}
A KG defines a set of common attributes, e.g., \textit{rdfs:label} and \textit{schema:description}, to describe its entities.
We encode the attribute triples in the KG and generate the attribute representations for corresponding entities.
For each attribute triple of an entity, we concatenate the attribute name $a$ and attribute value $v$ into a token sequence $q=[a;v]=(w_1,\dots,$ $w_M)$. 
In order to cope with the out-of-vocabulary problem, we define a lookup function to convert each token to a token embedding:
\begin{align}
    \mathrm{LP}(w_j) = \left\{
	\begin{array}{ll}
		\mathrm{WordEmb}(w_j), & \text{if } w_j \text{ has word emb.}\\
		\mathrm{CharEmb}(w_j), & \text{otherwise}
	\end{array},
	\right.
\end{align}
where $\mathrm{WordEmb (\cdot)}$ returns the word embedding in GloVe, and $\mathrm{CharEmb (\cdot)}$ offers the average of character embeddings pre-trained with Skip-gram.
Our method can work with other word or character embeddings easily. 

Next, we leverage AutoEncoder to encode a sequence of token embeddings into an attribute triple embedding in an unsupervised way:
\begin{align}
	\mathbf{q} = \textrm{AutoEncoder} \Big(\big[\textrm{LP} (w_1);\dots;\textrm{LP} (w_M)\big]\Big), 
\end{align}
where AutoEncoder is pre-trained on the attribute triples. 
We conduct self-supervised training, and both encoder and decoder of AutoEncoder use BiLSTM. 
AutoEncoder has good capacity for feature extraction and compression. In our model, the input of AutoEncoder is a concatenation vector of an entity and its attributes. The reconstruction loss of AutoEncoder can help extract a better compressed feature representation while preserving the attribute knowledge.

Finally, we stack all attribute triple embeddings of an entity into a one-dimensional CNN to obtain the attribute representation of the entity:
\begin{align}
    \mathbf{h}_{e_i}^{0} = \mathrm{MaxPooling} \big(\mathrm{CNN_{1D}} (\parallel_j \mathbf{q}_j)\big),
\end{align}
where $\parallel$ denotes the stack operation, and $\mathbf{h}_{e_i}^{0}$ is the attribute representation of entity $e_i$, which would be used as the input representation for relation triple encoding below. Here, we choose CNN since the convolutional layer is a good feature extractor to learn high-level representations from value embeddings while reducing the dimension of output representations. 
Furthermore, we use the 1D convolution kernel as its invariance to the order of attribute embeddings. 

\textbf{Relation triple encoding.}
The relation triples present in the form of an entity-relation graph structure, and the topology and relation types are the key to encode such knowledge.
Based on the attribute representations of entities, we employ a R-GAT \cite{busbrige2019relational} with $K$ layers to convolute the entity-relation graph. 
R-GAT incorporates relation types using different embeddings and calculates attention scores on all adjacent nodes based on entity embeddings and relation embeddings.
Specifically, the node forward-pass update for the $(k+1)^\text{th}$ layer is
\begin{align}
\begin{aligned}
	\mathbf{e}_{ij}^{(k,b)} &= \mathbf{W}_\text{out}^{(k,b)^{T}} \big[\mathbf{W}_\text{in}^{(k,b)} \mathbf{h}_i^{(k)} ; \mathbf{W}_\text{in}^{(k,b)} \mathbf{h}_j^{(k)} ; \mathbf{M}(r_{ij})\big],\\
	\alpha_{ij}^{(k,b)} &= \frac{\exp \big(\mathrm{LeakyReLU} ( \mathbf{e}_{ij}^{(k,b)})\big)} {\sum_{l \in U_i} \exp \big(\mathrm{LeakyReLU} ( \mathbf{e}_{il}^{(k,b)})\big)}, \\
	\mathbf{h}_i^{(k+1)} &= \frac{1}{B}\sum_{b=1}^B \sigma \Big(\sum_{l \in U_i} \alpha_{il}^{(k,b)} \mathbf{W}_\text{in}^{(k,b)} \mathbf{h}_l^{(k)}\Big),
\end{aligned}
\end{align}
where $\mathbf{W}_\text{in}^{(k, b)}$ and $\mathbf{W}_\text{out}^{(k,b)}$ denote two trainable parameters of the $b^\text{th}$ attention head ($1\le b\le B$) at the $k^\text{th}$ layer. 
$\mathbf{h}_i^{(k)}$ and $\mathbf{h}_j^{(k)}$ are the node representations of entities $e_i$ and $e_j$ at the $k^\text{th}$ layer, respectively. 
$\mathbf{M}$ is a trainable mapping matrix corresponding to the relation types in the KG. 
$r_{ij}$ is the relation type between $e_i$ and $e_j$. 
$\mathrm{LeakyReLU}(\cdot)$ and $\sigma(\cdot)$ are the activation functions. 
$U_i$ is the neighbor set of $e_i$. In this way, the entity representations are updated via their all adjacent entity embeddings and relation embeddings at the previous layer.

We refer to the representations of entities after graph convolution as the \emph{KG representations} of entities, which encode the knowledge in both attribute triples and relation triples. 
We simply denote the KG representation of entity $e_i$ by $\mathbf{h}_{e_i}$.

In the KG encoding, the attribute encoding has $d_\text{Auto}N_\text{max}N_\text{kernel}(d_\text{kernel}^2+1)$ parameters, where $d_\text{Auto}$ is the output dimension of AutoEncoder, $N_\text{max}$ is the maximum number of attributes that an entity has, $N_\text{kernel}$ is the number of kernels, and $d_\text{kernel}$ is the kernel size of CNN. 
The R-GAT network in the relation triple encoding contains $2(N_\text{layer}-1)N_\text{head}d_\text{RGAT}^2+d_\text{RGAT}d_\text{ent}$ parameters, where $N_\text{layer}$ is the number of layers in R-GAT, $N_\text{head}$ is the number of attention heads at each layer of R-GAT, $d_\text{RGAT}$ is the hidden dimension of R-GAT hidden layers, and $d_\text{ent}$ is the dimension of entity representations.
\subsection{Representation Reconciliation}
\label{subsect:aggr}

The token representations and the KG representations of entities capture different knowledge in independent semantic spaces. 
Following ERNIE \cite{zhang2019ernie} and K-BERT \cite{liu2020kbert}, we employ a representation reconciliation module to exchange the knowledge from entities in the KG with their linked tokens in the document.

The representation reconciliation module consists of $N$-stacked aggregators. 
For the $n^\text{th}$ ($1\le n\le N$) aggregator, given the token representations $\{\mathbf{h}_{w_1}^{n-1},\dots,$ $\mathbf{h}_{w_J}^{n-1}\}$ and the KG representations of entities $\{\mathbf{h}_{e_1}^{n-1}, \dots, \mathbf{h}_{e_I}^{n-1}\}$ from the preceding aggregator, the fusion phase is formulated as
\begin{align}
\tilde{\mathbf{h}} = \left\{
\begin{array}{ll}
	\sigma (\tilde{\mathbf{W}}_w^{(n)} \tilde{\mathbf{h}}_{w_j}^{(n)} + \tilde{\mathbf{W}}_e^{(n)} \tilde{\mathbf{h}}_{e_i}^{(n)} + \tilde{\mathbf{b}}^{(n)}), & \quad\text{if } w_j, e_i \text{ align} \\
	\sigma (\tilde{\mathbf{W}}_w^{(n)} \tilde{\mathbf{h}}_{w_j}^{(n)} + \tilde{\mathbf{b}}^{(n)}), & \quad\text{otherwise} \\
\end{array},
\right.
\end{align}
where $\tilde{\mathbf{h}}_{w_j}^{(n)},\tilde{\mathbf{h}}_{e_i}^{(n)}$ are the token representation and KG representation after the multi-head self-attention \cite{vaswani2017attention}, respectively. 
$\tilde{\mathbf{W}}_w^{(n)}, \tilde{\mathbf{W}}_e^{(n)},\tilde{\mathbf{b}}^{(n)}$ are three trainable parameters.
The information in the two semantic spaces is mutually integrated.

Then, the reconstruction phase leverages $\tilde{\mathbf{h}}$ to refine the output representations of each token and entity in the aligned token-entity pairs:
\begin{align}
\begin{aligned}
	\mathbf{h}_{w_j}^{(n)} &= \sigma (\mathbf{W}_w^{(n)} \tilde{\mathbf{h}} + \mathbf{b}_w^{(n)}), \\
	\mathbf{h}_{e_i}^{(n)} &= \sigma (\mathbf{W}_{e}^{(n)} \tilde{\mathbf{h}} + \mathbf{b}_e^{(n)}).
\end{aligned}
\end{align}
Here, the aligned token representations and entity representations are updated and enhanced by the integrated information.
Note that the representations of entities without aligned tokens would not be updated.

Finally, we obtain the token representation sequence $\{\mathbf{h}_{w_1}^{N}, \dots, \mathbf{h}_{w_J}^{N}\}$ from the last aggregator, which would constitute $\mathbf{H}'$ and be fed to the prediction layer. 

To supervise the above process, we employ a token-entity alignment task. 
For each aligned token-entity pair $(w_j, e_i)$, we predict the aligned KG entity $e_i$ based on the token $w_j$. 
We only ask the model to predict entities within a given entity candidate set. 
By default, all linked entities in the document form the candidate set. 
For the token sequence $\{ w_1,$ $\dots, w_J\}$ and the corresponding candidate entities $\{e_1,\dots,$ $e_I\}$, the token-entity alignment loss is
\begin{align}
    \mathcal{L}_{kg} = \sum_{j=1}^{J} \sum_{i=1}^{I} f_{j,i}^* * P(e_i\,|\,w_j),
\end{align}
where $f_{j,i}^*\in\{0,1\}$ is the true alignment label between $w_j$ and $e_i$, and $P (e_i\,|\,w_j) = \frac{\mathrm{exp} \big(\textrm{Linear} (\mathbf{h}_{w_j}^{(N)}) \cdot \mathbf{h}_{e_i}\big)} {\sum_{l=1}^{I} \mathrm{exp} \big(\mathrm{Linear} (\mathbf{h}_{w_j}^{(N)}) \cdot \mathbf{h}_{e_l}\big)}$ returns the probability that $e_i$ can be predicted by $w_j$.

We optimize the RE loss, coreference loss and token-entity alignment loss with multi-task learning. 
The final loss is
\begin{align}
    \mathcal{L} = \alpha_1\cdot \mathcal{L}_{re} + \alpha_2\cdot \mathcal{L}_{cr} + \alpha_3\cdot \mathcal{L}_{kg},
\end{align}
where $\alpha_1, \alpha_2$ and $\alpha_3$ are the weight hyperparameters.

In the representation reconciliation, for each aggregator, the multi-head self-attention networks contain $4d_\text{token}^2+4d_\text{ent}^2$ parameters, the fusion phase contains $d_\text{out}(N_\text{token}+N_\text{align})+N_\text{token}$ parameters, and the reconstruction phase contains $2N_\text{align}(d_\text{out}+1)$ parameters, where $d_\text{out}$ is the output dimension of multi-head self-attention networks, $N_\text{token}$ is the number of tokens, and $N_\text{align}$ is the number of aligned token-entity pairs. 
Therefore, the parameters of representation reconciliation are $N_\text{agg}\big[4d_\text{token}^2+4d_\text{ent}^2+d_\text{out}(N_\text{token}+N_\text{align})+N_\text{token}+2N_\text{align}(d_\text{out}+1)\big]$, where $N_\text{agg}$ is the number of aggregators.

\textbf{Model complexity.}
The total parameter number of KIRE is $d_\text{MLP}(2d_\text{token}+d_\text{dist})+2d_\text{token}+d_\text{Auto}N_\text{max}(d_\text{kernel}^2+1)+2(N_\text{layer}-1)N_\text{head}d_\text{RGAT}^2+d_\text{RGAT}d_\text{ent} + N_\text{agg}\big[4d_\text{token}^2+4d_\text{ent}^2 +d_\text{out}(N_\text{token}+N_\text{align})+N_\text{token}+2N_\text{align}(d_\text{out}+1)\big].$
\section{Experiments and Results}
\label{sect:exp}

We develop KIRE with PyTorch 1.7.1, and test on an X86 server with two Xeon Gold 5117 CPUs, 250 GB memory, two Titan RTX GPUs and Ubuntu 18.04. 

\subsection{Experiment Setup}
\label{subsect:setup}

\textbf{Datasets.} 
We select two benchmark datasets in our experiments:
(1) DocRED \cite{yao2019docred} is a crowdsourced dataset for document-level RE.
The relation labels in its test set are not public.
(2) DWIE \cite{zaporojets2021dwie} is a new dataset for document-level multi-task information extraction. 
We use the data relevant to RE only.
Since DWIE does not have the validation set, we randomly split its training set into 80\% for training and 20\% for validation. Table~\ref{tab:stat} lists the statistical data.

\begin{table}[!tb]\setlength\tabcolsep{2pt}
\caption{Dataset statistics. Inst. denotes relation instances excluding N/A relation.}
\centering
\resizebox{.65\columnwidth}{!}{
\begin{tabular}{l|l|rcrr}
\hline 
\multicolumn{2}{c|}{Datasets} & \#Doc. & \#Rel. & \#Inst. & \#N/A Inst. \\ 
\hline  
\multirow{3}{*}{DocRED} & Training set & 3,053 & 96 & 38,269 & 1,163,035 \\
			~ & Validation set & 1,000 & 96 & 12,332 & 385,263 \\
			~ & Test set & 1,000 & 96 & 12,842 & 379,316 \\
\hline  
\multirow{3}{*}{DWIE} & Training set & 544 & 66 & 13,524 & 492,057 \\
			~ & Validation set & 137 & 66 & 3,488 & 121,750\\
			~ & Test set & 96 & 66 & 2,453 & 78,995 \\
\hline
\end{tabular}}
\label{tab:stat} 
\end{table}

\textbf{Knowledge graph.} We select Wikidata (2020-12-01) as our KG due to its coverage and popularity \cite{bastos2021recon,vashishth2018reside}. 
The numbers of its relation and attribute triples are 506,809,195 and 729,798,070, respectively. 
To prevent the test leakage, we filter out all the relation triples with entity pairs to be labeled in the test sets. 

\textbf{Evaluation metrics.} 
We measure F1-score (F1) and Ignore F1-score (Ign F1) in our experiments. 
We repeat five times using the same hyperparameters but different random seeds, and report the means and standard deviations.

\textbf{Implementation details.} To achieve good generalization, we do not carry out excessive feature engineering on Wikidata. 
Numerical attributes are regarded as texts, and their semantics are captured by the word embeddings \cite{pennington2014glove}.
We employ NeuralCoref 4.0 as our coreference resolution tool and also use the annotations provided in DocRED and DWIE.
We use a two-stage method for training, which first trains a basic RE model and then fine-tunes this model to train the knowledge injection layer.
The training procedure is optimized with Adam.
Moreover, to compare fairly, the basic RE model and its corresponding KIRE adopt the same hyperparameter values. 
We set the batch size to 4 and the learning rate to 0.0005. 
We use three R-GAT layers and two aggregators.
Moreover, $\alpha_1,\alpha_2,\alpha_3$ are 1, 0.01 and 0.01, respectively. 
The dimension of hidden layers in MLP is 256, the dimensions of GloVe and Skip-gram are 100, and the dimension of hidden layers in AutoEncoder is 50. 
See the source code for more details.

\begin{table}[!tb]\setlength\tabcolsep{2pt}
\caption{Comparison of result improvement on baseline models}
\label{tab:basic}
\centering
\resizebox{\columnwidth}{!}{
	\begin{tabular}{l|ccccccccccc}
		\hline 
		\multirow{3}{*}{Models} & \multicolumn{5}{c}{DocRED} & & \multicolumn{5}{c}{DWIE} \\ 
		\cline{2-6} \cline{8-12} & \multicolumn{2}{c}{Validation set} & & \multicolumn{2}{c}{Test set} & &  \multicolumn{2}{c}{Validation set} & & \multicolumn{2}{c}{Test set} \\
		\cline{2-3} \cline{5-6} \cline{8-9} \cline{11-12} & Ign F1 & F1 & & Ign F1 & F1 & & Ign F1 & F1 & & Ign F1 & F1 \\
		\hline 
		CNN                 & $43.91_{\pm 0.02}$ & $45.99_{\pm 0.05}$ & & $42.61_{\pm 0.06}$ & $44.80_{\pm 0.08}$ & & $38.21_{\pm 0.04}$ & $49.09_{\pm 0.06}$ & & $40.06_{\pm 0.08}$ & $51.21_{\pm 0.13}$ \\
		\ +\,KIRE        & $\mathbf{46.18}_{\pm 0.04}$ & $\mathbf{48.21}_{\pm 0.06}$ & & $\mathbf{45.24}_{\pm 0.05}$ & $\mathbf{47.27}_{\pm 0.08}$ & & $\mathbf{39.68}_{\pm 0.05}$ & $\mathbf{50.49}_{\pm 0.09}$ & & $\mathbf{42.09}_{\pm 0.04}$ & $\mathbf{53.16}_{\pm 0.08}$ \\
		\ +\,RESIDE      & $45.03_{\pm 0.06}$ & $47.12_{\pm 0.08}$ & & $43.79_{\pm 0.05}$ & $45.96_{\pm 0.09}$ & & $39.24_{\pm 0.04}$ & $50.13_{\pm 0.07}$ & & $41.31_{\pm 0.06}$ & $52.47_{\pm 0.11}$ \\
		\ +\,RECON       & $45.57_{\pm 0.04}$ & $47.64_{\pm 0.07}$ & & $44.53_{\pm 0.07}$ & $46.68_{\pm 0.10}$ & & $39.42_{\pm 0.03}$ & $50.34_{\pm 0.06}$ & & $41.73_{\pm 0.07}$ & $52.74_{\pm 0.09}$ \\
		\ +\,KB-graph    & $45.49_{\pm 0.03}$ & $47.58_{\pm 0.08}$ & & $44.46_{\pm 0.06}$ & $46.61_{\pm 0.09}$ & & $39.34_{\pm 0.06}$ & $50.26_{\pm 0.09}$ & & $41.65_{\pm 0.08}$ & $52.63_{\pm 0.12}$ \\
		\hline 
		LSTM                & $48.49_{\pm 0.05}$ & $50.41_{\pm 0.07}$ & & $47.41_{\pm 0.04}$ & $49.47_{\pm 0.10}$ & & $52.79_{\pm 0.03}$ & $63.61_{\pm 0.08}$ & & $54.87_{\pm 0.07}$ & $65.17_{\pm 0.14}$ \\
		\ +\,KIRE        & $\mathbf{50.41}_{\pm 0.03}$ & $\mathbf{52.49}_{\pm 0.06}$ & & $\mathbf{49.55}_{\pm 0.06}$ & $\mathbf{51.72}_{\pm 0.09}$ & & $\mathbf{54.11}_{\pm 0.04}$ & $\mathbf{64.86}_{\pm 0.08}$ & & $\mathbf{56.74}_{\pm 0.05}$ & $\mathbf{66.91}_{\pm 0.07}$ \\
		\ +\,RESIDE      & $49.58_{\pm 0.04}$ & $51.49_{\pm 0.08}$ & & $48.52_{\pm 0.06}$ & $50.51_{\pm 0.09}$ & & $53.87_{\pm 0.02}$ & $64.56_{\pm 0.06}$ & & $55.96_{\pm 0.06}$ & $66.29_{\pm 0.12}$ \\
		\ +\,RECON       & $50.03_{\pm 0.03}$ & $51.98_{\pm 0.08}$ & & $49.07_{\pm 0.07}$ & $51.12_{\pm 0.12}$ & & $53.98_{\pm 0.03}$ & $64.69_{\pm 0.07}$ & & $56.35_{\pm 0.04}$ & $66.51_{\pm 0.08}$ \\
		\ +\,KB-graph    & $49.94_{\pm 0.04}$ & $51.89_{\pm 0.07}$ & & $48.98_{\pm 0.05}$ & $51.04_{\pm 0.09}$ & & $53.91_{\pm 0.05}$ & $64.61_{\pm 0.08}$ & & $56.27_{\pm 0.06}$ & $66.43_{\pm 0.09}$ \\
		\hline 
		BiLSTM              & $48.51_{\pm 0.04}$ & $50.54_{\pm 0.08}$ & & $47.58_{\pm 0.05}$ & $49.66_{\pm 0.11}$ & & $53.95_{\pm 0.05}$ & $63.96_{\pm 0.07}$ & & $54.91_{\pm 0.09}$ & $65.39_{\pm 0.11}$ \\
		\ +\,KIRE        & $\mathbf{50.46}_{\pm 0.02}$ & $\mathbf{52.65}_{\pm 0.05}$ & & $\mathbf{49.69}_{\pm 0.04}$ & $\mathbf{51.98}_{\pm 0.07}$ & & $\mathbf{55.86}_{\pm 0.05}$ & $\mathbf{65.77}_{\pm 0.09}$ & & $\mathbf{56.88}_{\pm 0.05}$ & $\mathbf{67.02}_{\pm 0.08}$ \\
		\ +\,RESIDE      & $49.64_{\pm 0.03}$ & $51.59_{\pm 0.06}$ & & $48.62_{\pm 0.04}$ & $50.71_{\pm 0.10}$ & & $55.04_{\pm 0.06}$ & $65.01_{\pm 0.09}$ & & $56.16_{\pm 0.05}$ & $66.47_{\pm 0.12}$ \\
		\ +\,RECON       & $49.97_{\pm 0.04}$ & $52.06_{\pm 0.07}$ & & $49.14_{\pm 0.06}$ & $51.32_{\pm 0.09}$ & & $55.42_{\pm 0.04}$ & $65.38_{\pm 0.08}$ & & $56.51_{\pm 0.06}$ & $66.63_{\pm 0.09}$ \\
		\ +\,KB-graph    & $49.89_{\pm 0.03}$ & $51.98_{\pm 0.07}$ & & $49.05_{\pm 0.05}$ & $51.26_{\pm 0.08}$ & & $55.35_{\pm 0.03}$ & $65.31_{\pm 0.09}$ & & $56.42_{\pm 0.07}$ & $66.55_{\pm 0.11}$ \\
		\hline 
		Context-aware       & $49.79_{\pm 0.03}$ & $51.84_{\pm 0.04}$ & & $48.73_{\pm 0.07}$ & $50.91_{\pm 0.12}$ & & $54.68_{\pm 0.04}$ & $64.29_{\pm 0.06}$ & & $56.53_{\pm 0.07}$ & $65.91_{\pm 0.09}$ \\
		\ +\,KIRE        & $\mathbf{51.07}_{\pm 0.03}$ & $\mathbf{53.25}_{\pm 0.07}$ & & $\mathbf{50.43}_{\pm 0.05}$ & $\mathbf{52.75}_{\pm 0.10}$ & & $\mathbf{56.58}_{\pm 0.03}$ & $\mathbf{65.62}_{\pm 0.07}$ & & $\mathbf{58.41}_{\pm 0.04}$ & $\mathbf{67.37}_{\pm 0.08}$ \\
		\ +\,RESIDE      & $50.43_{\pm 0.04}$ & $52.59_{\pm 0.07}$ & & $49.58_{\pm 0.05}$ & $51.86_{\pm 0.09}$ & & $55.74_{\pm 0.03}$ & $65.11_{\pm 0.07}$ & & $57.64_{\pm 0.05}$ & $66.78_{\pm 0.08}$ \\
		\ +\,RECON       & $50.78_{\pm 0.03}$ & $52.89_{\pm 0.06}$ & & $49.97_{\pm 0.04}$ & $52.27_{\pm 0.08}$ & & $56.12_{\pm 0.05}$ & $65.48_{\pm 0.08}$ & & $58.02_{\pm 0.06}$ & $66.94_{\pm 0.10}$ \\
		\ +\,KB-graph    & $50.69_{\pm 0.05}$ & $52.81_{\pm 0.07}$ & & $49.88_{\pm 0.06}$ & $52.19_{\pm 0.11}$ & & $56.03_{\pm 0.04}$ & $65.39_{\pm 0.09}$ & & $57.94_{\pm 0.05}$ & $66.89_{\pm 0.11}$ \\
		\hline
	\end{tabular}}
\end{table}

\subsection{Main Results}
\label{subsect:main}

\textbf{Improvement on baseline models.}
To validate the effectiveness and versatility of KIRE, we pick four baseline models in \cite{yao2019docred}.
The first three models directly employ CNN, LSTM and BiLSTM to encode documents, while the fourth model is called context-aware, which leverages the attention mechanism with BiLSTM. 
These four models are native to the DocRED dataset and widely chosen as the competitors in many RE studies \cite{huang2021three,nan2020reasoning,tang2020hin,wang2020global,wang2019fine,xu2021entity,zeng2020double,zhou2021document}.

Table~\ref{tab:basic} depicts the result improvement, and we observe that: 
(1) KIRE consistently improves the performance of all baselines on DocRED and DWIE, which demonstrates the good generalization of KIRE. 
Small standard deviations also tells the good stability of KIRE.
(2) KIRE obtains a significant improvement of Ign F1/F1 up to 2.63/2.47 on DocRED and 2.03/1.95 on DWIE, respectively. 
This is mainly because the ways that the baseline models encode a document are too simple to capture some part of important contextual information in the document. 
External knowledge from KIRE makes up for this part, and therefore effectively improves the model performance.
(3) CNN performs poorly, because the text order is important for RE while CNN cannot process such order well. 

\textbf{Comparison with existing knowledge injection models.}
We choose three recent models: RESIDE \cite{vashishth2018reside}, RECON \cite{bastos2021recon} and KB-graph \cite{verlinden2021injecting}, which inject extra knowledge into RE models.
Specifically, we use KB-graph instead of the full version KB-both since it selects Wikipedia as another knowledge source, which is unfair to other models.
To compare fairly, we only adopt the knowledge injection modules of the above models to enhance the token representations in the documents, and the representations are used by the baseline RE models to predict the relation labels.

Table~\ref{tab:basic} presents the comparison results, and we obtain several findings: 
(1) KIRE is consistently superior to RESIDE, RECON and KB-graph with an improvement of Ign F1/F1 up to 0.71/0.66 on DocRED and 0.39/0.43 on DWIE, respectively. 
Given that the test sets contain (ten) thousand relation instances, we think that the improvement makes sense.
For example, on the validation set of DocRED, KIRE can correctly predict an average of 478 more instances than the second best method RECON.
Such improvement brought by KIRE attributes to that KIRE absorbs more knowledge like coreferences and fuses the knowledge better.
(2) The improvement brought by RESIDE is the lowest since it only injects limited knowledge like entity types and relation aliases. 
RECON and KB-graph explore more knowledge from the KG, but they still ignore the coreference knowledge. 
Besides, the methods that they employ to integrate knowledge are representation average or concatenation, which may lose part of semantic information in the injected knowledge.

\begin{table}[!tb]\setlength\tabcolsep{2pt}
\caption{Result improvement on state-of-the-art models}
\label{tab:complex}
\centering
\resizebox{\columnwidth}{!}{
	\begin{tabular}{l|ccccccccccc}
		\hline
		\multirow{3}{*}{Models} & \multicolumn{5}{c}{DocRED} & & \multicolumn{5}{c}{DWIE} \\ 
		\cline{2-6} \cline{8-12} & \multicolumn{2}{c}{Validation set} & & \multicolumn{2}{c}{Test set} & &  \multicolumn{2}{c}{Validation set} & & \multicolumn{2}{c}{Test set} \\
		\cline{2-3} \cline{5-6} \cline{8-9} \cline{11-12} & Ign F1 & F1 & & Ign F1 & F1 & & Ign F1 & F1 & & Ign F1 & F1 \\
		\hline
		ATLOP           & $59.25_{\pm 0.03}$ & $61.14_{\pm 0.07}$ & & $58.32_{\pm 0.05}$ & $60.44_{\pm 0.08}$ & & $69.12_{\pm 0.04}$ & $76.32_{\pm 0.09}$ & & $73.85_{\pm 0.08}$ & $80.38_{\pm 0.12}$ \\
		\ +\,KIRE    & $\mathbf{59.58}_{\pm 0.04}$ & $\mathbf{61.45}_{\pm 0.09}$ & & $\mathbf{59.35}_{\pm 0.06}$ & $\mathbf{61.39}_{\pm 0.11}$ & & $\mathbf{69.75}_{\pm 0.05}$ & $\mathbf{76.75}_{\pm 0.08}$ & & $\mathbf{74.43}_{\pm 0.07}$ & $\mathbf{80.73}_{\pm 0.15}$ \\
		\hline
		SSAN            & $56.68_{\pm 0.03}$ & $58.95_{\pm 0.04}$ & & $56.06_{\pm 0.05}$ & $58.41_{\pm 0.06}$ & & $51.80_{\pm 0.05}$ & $62.87_{\pm 0.10}$ & & $57.49_{\pm 0.09}$ & $67.77_{\pm 0.12}$ \\
		\ +\,KIRE    & $57.29_{\pm 0.05}$ & $59.31_{\pm 0.06}$ & & $56.31_{\pm 0.06}$ & $58.65_{\pm 0.08}$ & & $52.67_{\pm 0.06}$ & $63.64_{\pm 0.10}$ & & $60.57_{\pm 0.09}$ & $69.58_{\pm 0.12}$ \\
        \hline
        GLRE            & $56.57_{\pm 0.06}$ & $58.43_{\pm 0.09}$ & & $55.40_{\pm 0.07}$ & $57.40_{\pm 0.13}$ & & $63.11_{\pm 0.03}$ & $71.21_{\pm 0.06}$ & & $62.95_{\pm 0.05}$ & $72.24_{\pm 0.09}$ \\
		\ +\,KIRE    & $57.31_{\pm 0.05}$ & $59.45_{\pm 0.10}$ & & $56.54_{\pm 0.09}$ & $58.49_{\pm 0.14}$ & & $65.17_{\pm 0.05}$ & $71.68_{\pm 0.09}$ & & $64.32_{\pm 0.06}$ & $73.35_{\pm 0.11}$ \\
		\hline
	\end{tabular}}
\end{table}

\textbf{Improvement on state-of-the-art models.}
We employ two sequence-based models, ATLOP \cite{zhou2021document} and SSAN \cite{xu2021entity}, as well as a graph-based model, GLRE~\cite{wang2020global}, due to their good performance and open source.
Enhancing these models is very challenging, since they have already explored various information in the documents and achieved state-of-the-art results. 
Due to the limit of GPU RAM, we use the BERT-base versions of ATLOP, SSAN and GLRE and re-run them according to the hyperparameters reported in their papers and source code.

The result improvement is shown in Table~\ref{tab:complex}, and we have several findings: 
(1) For the two sequence-based models, KIRE obtains an improvement of Ign F1/F1 up to 1.03/0.95 on DocRED and 3.08/1.81 on DWIE, respectively. 
This mainly attributes to the fact that the extra knowledge injected by KIRE can effectively help the models identify and capture more interactions between entity pairs especially across sentences.
(2) For the graph-based model, KIRE obtains an improvement of Ign F1/F1 up to 1.14/1.09 on DocRED and 1.37/1.11, respectively. 
This is largely due to the fact that the extra knowledge injected by KIRE can enrich the representations of mention nodes and entity nodes in the document graphs for more accurate reasoning between entity pairs especially of longer distance.
(3) This also verifies that our knowledge injection framework can be generalized to a broad range of document-level RE models.



\subsection{Detailed Analysis}
\label{subsect:analysis}

\textbf{Ablation study.} We conduct an ablation study on the four baseline models. 
For ``w/o distill'', we disable the coreference distillation module and directly use the original coreferences as the injected knowledge. 
For ``w/o attr.'', we initialize the relation triple representations by max pooling the word embeddings of entity labels.
For ``w/o rel.'', we directly adopt the attribute representations of entities as KG representations. 
For ``w/o KG'', we disable the whole KG encoding module.
Additionally, we replace KIRE with three simple variants for knowledge injection.
For ``w/ rep. avg'', we average the hidden representations of alias mentions, and the token representations are averaged with KG representations of entities.
For ``w/ rep. concat'', we concatenate the representations of alias mentions, and the KG representations of entities are concatenated after the aligned token representations.
For ``w/ MLP'', we leverage two MLP layers to fuse the representations of alias mentions and the KG representations of entities with the aligned token representations, respectively.

\begin{figure}[!tb]
	\centering
	\includegraphics[width=\columnwidth]{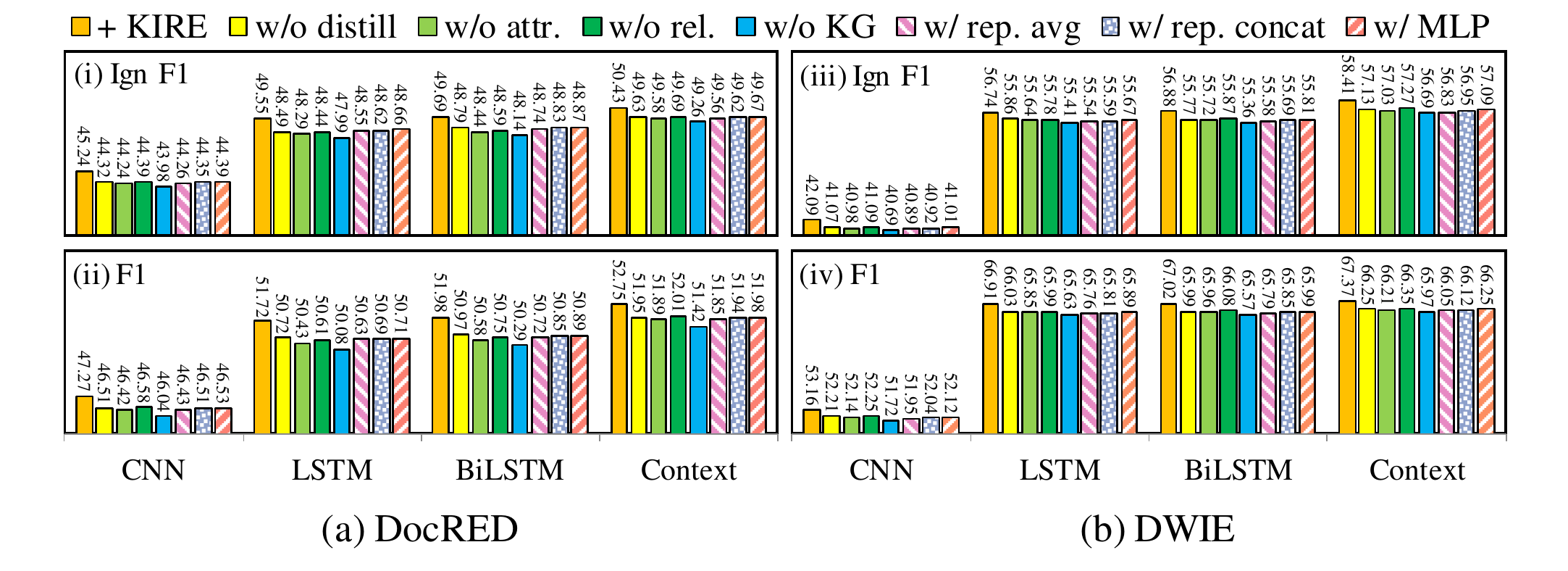}
	\caption{Results of ablation study}
	\label{fig:ablation}
\end{figure}

From Figure~\ref{fig:ablation}, we can see that: 
(1) Ign F1/F1-scores reduce when we disable any modules, showing their contributions. 
(2) The changes caused by removing one type of knowledge are not obvious, mainly due to the crossovers among the three types of knowledge in the information space. 
(3) The results decline if we disable the coreference distillation, due to the coreference errors in the injected knowledge.
(4) If we remove the KG encoding, the results drop drastically, as the baseline models cannot generate extra relation and attribute knowledge. 
(5) Compared to the three variants, the larger increase brought by KIRE validates the effectiveness of coreference distillation and representation reconciliation.

\textbf{Influence of mention number.} We measure the effectiveness of KIRE w.r.t. average mention number for each entity pair. 
For DocRED, we evaluate it on the validation set.
The results are shown in Table~\ref{tab:number}. 
We observe that KIRE gains higher performance for the entity pairs with more mentions, in particular when the average mention number $> 3$.
This is because KIRE injects knowledge into the RE models by updating the token representations of entity mentions, which has a greater impact on the entities with more mentions.

\begin{table}[!tb]\setlength\tabcolsep{2pt}
\caption{Results w.r.t. average mention number}
\label{tab:number}
\centering
\resizebox{\columnwidth}{!}{
	\begin{tabular}{l|ccccccccccccccccc}
		\hline 
		\multirow{3}{*}{Models} & \multicolumn{8}{c}{DocRED} & & \multicolumn{8}{c}{DWIE} \\ 
		\cline{2-9} \cline{11-18} & \multicolumn{2}{c}{$1$} & & \multicolumn{2}{c}{$(1,3]$} & &  \multicolumn{2}{c}{$>3$} & & \multicolumn{2}{c}{$1$} & & \multicolumn{2}{c}{$(1,3]$} & &  \multicolumn{2}{c}{$>3$} \\
		\cline{2-3} \cline{5-6} \cline{8-9} \cline{11-12} \cline{14-15} \cline{17-18} & Ign F1 & F1 & & Ign F1 & F1 & & Ign F1 & F1 & & Ign F1 & F1 & & Ign F1 & F1 & & Ign F1 & F1\\
		\hline 
		CNN             & 42.24 & 44.27 & & 44.42 & 46.45 & & 45.38 & 47.35 & & 39.35 & 50.44 & & 40.81 & 51.96 & & 41.94 & 53.05 \\
		\ +\,KIRE    & $\uparrow$1.64 & $\uparrow$1.75 & & $\uparrow$2.71 & $\uparrow$2.89 & & $\uparrow$3.21 & $\uparrow$3.43 & & $\uparrow$1.52 & $\uparrow$1.49 & & $\uparrow$2.35 & $\uparrow$2.32 & & $\uparrow$2.95 & $\uparrow$2.98 \\
		\hline
		LSTM            & 47.39 & 49.35 & & 48.59 & 50.53 & & 50.09 & 52.08 & & 54.14 & 64.41 & & 55.97 & 66.14 & & 57.21 & 67.55 \\
		\ +\,KIRE    & $\uparrow$1.76 & $\uparrow$1.87 & & $\uparrow$2.65 & $\uparrow$2.82 & & $\uparrow$3.37 & $\uparrow$3.51 & & $\uparrow$1.31 & $\uparrow$1.26 & & $\uparrow$1.87 & $\uparrow$1.89 & & $\uparrow$2.35 & $\uparrow$2.29 \\
		\hline
		BiLSTM          & 47.35 & 49.32 & & 48.61 & 50.54 & & 50.21 & 52.28 & & 54.07 & 64.61 & & 56.01 & 66.25 & & 57.30 & 67.79 \\
		\ +\,KIRE    & $\uparrow$1.83 & $\uparrow$1.95 & & $\uparrow$2.73 & $\uparrow$2.91 & & $\uparrow$3.31 & $\uparrow$3.43 & & $\uparrow$1.49 & $\uparrow$1.12 & & $\uparrow$1.91 & $\uparrow$1.86 & & $\uparrow$2.34 & $\uparrow$2.16 \\
		\hline
		Context-aware   & 48.33 & 50.19 & & 49.63 & 51.64 & & 51.10 & 53.34 & & 55.98 & 65.16 & & 58.02 & 67.01 & & 58.72 & 68.41 \\
        \ +\,KIRE    & $\uparrow$1.43 & $\uparrow$1.56 & & $\uparrow$2.35 & $\uparrow$2.47 & & $\uparrow$2.87 & $\uparrow$2.98 & & $\uparrow$1.35 & $\uparrow$0.91 & & $\uparrow$1.92 & $\uparrow$1.57 & & $\uparrow$2.32 & $\uparrow$1.95 \\
		\hline
	\end{tabular}}
\end{table}

\begin{figure}[!tb]
	\centering
	\includegraphics[width=\columnwidth]{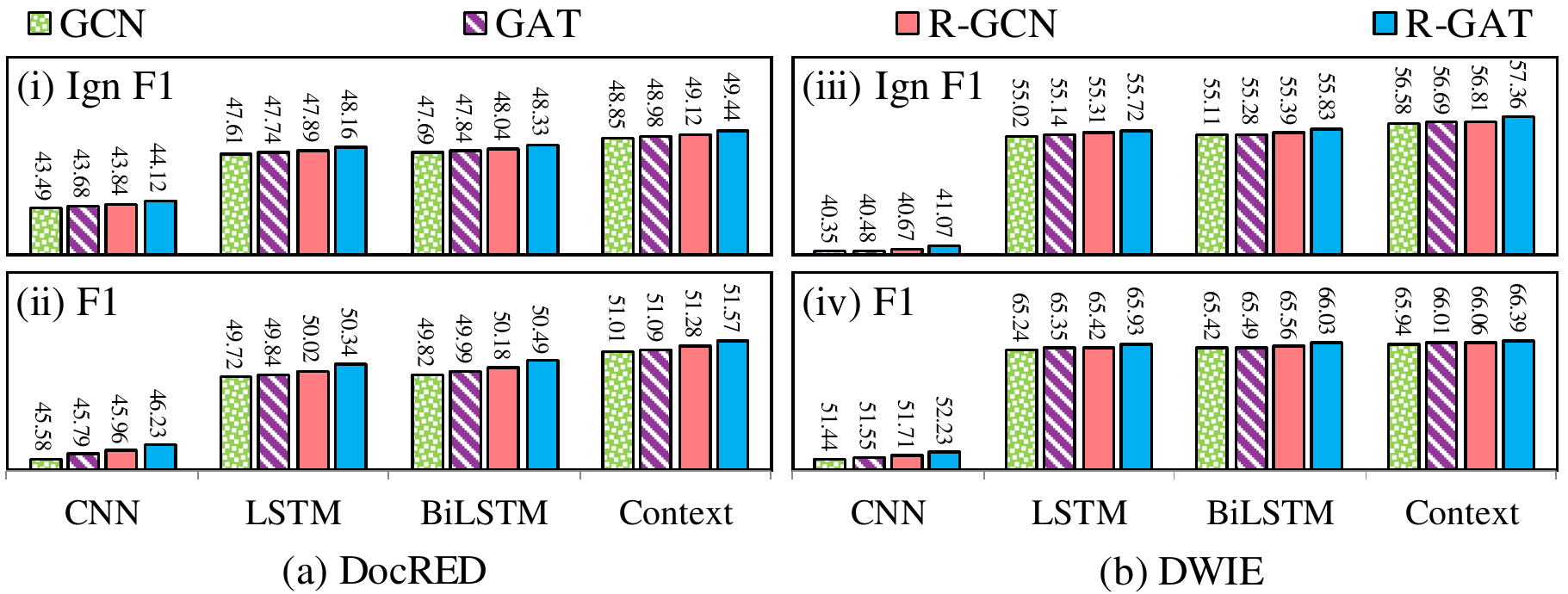}
	\caption{Result comparison of graph encoders}
	\label{fig:graph}
\end{figure}

\textbf{Comparison with alternative graph encoders.}
We compare R-GAT with GCN \cite{kipf2017semi}, GAT \cite{velivckovic2017graph} and R-GCN \cite{schlichtkrull2018modeling}. 
We remove the coreference encoding and attribute triple encoding to eliminate their interference.
From Figure~\ref{fig:graph}, the performance of R-GCN and R-GAT is better than GCN and GAT, as they can capture the relation information in the entity-relation graphs. 
The results of GAT are greater than GCN, as GAT can selectively aggregate the neighboring information by self-attention. 
Similarly, R-GAT slightly outperforms R-GCN.

\textbf{Case study.}  
We depict two successful cases and a failed case in Table \ref{tab:cases}. We still use CNN, LSTM, BiLSTM and Context-aware as baselines.
\begin{itemize}
\item \textbf{Case 1.} To identify the relation between \textit{North Dumfries} in $S2$ and \textit{Regional Municipality of Waterloo} in $S1$, we use the extra knowledge (\textit{Regional Municipality of Waterloo}, \textit{instance of}, \textit{regional municipality of Ontario}) and the coreference of \textit{It} and \textit{The Waterloo Moraine} from NeuralCoref to correctly infer the relation \textit{P131 (located in the administrative territorial entity)}.

\item \textbf{Case 2.} With the aid of the extra knowledge (\textit{Meghan Markle}, \textit{country of citizenship}, \textit{United States of America}) and the coreference of \textit{Meghan Markle} and \textit{Markle} from NeuralCoref, we successfully detect the relation \textit{citizen\_of} between \textit{Thomas Markle} in $S10$ and \textit{America} in $S1$.

\item \textbf{Case 3.} To recognize the relation between \textit{Ensign George Gay} in $S2$ and \textit{the United States Navy} in $S4$, we require a bridge entity \textit{Robert Kingsbury Huntington}. 
Through the coreference of \textit{Robert} and \textit{He} from NeuralCoref, we identify that \emph{Robert} and \textit{George} are comrades. 
Then from $S4$, we find out that \textit{Robert} is enlisted in \textit{the US Navy}. 
According to this reasoning chain, we can see that the relation between \textit{George} and \textit{the US Navy} is \textit{P241 (military branch)}. 
KIRE fails to run such complex reasoning involving three sentences.
\end{itemize}

\begin{table}[!tb]
\caption{Case study. \textit{\textbf{\textcolor{magenta}{Target entities}}} and \textit{\textbf{\textcolor{cyan}{related entities}}} are colored.}
\label{tab:cases} 
\centering
\resizebox{\columnwidth}{!}{
	\begin{tabular}{llll}
		\hline
		$[S1]$ & \multicolumn{3}{p{.95\columnwidth}}{\textit{\textbf{\textcolor{cyan}{The Waterloo Moraine}}} ... was created as a moraine in the \textit{\textbf{\textcolor{magenta}{Regional Municipality of Waterloo}}}, in Ontario, Canada.} \\
		$[S2]$ & \multicolumn{3}{p{.95\columnwidth}}{\textit{\textbf{\textcolor{cyan}{It}}} covers ... and some parts of the townships of Wellesley and \textit{\textbf{\textcolor{magenta}{North Dumfries}}}.} \\
		\hline
		\multicolumn{4}{p{1\columnwidth}}{\small \textbf{Case 1}\quad Gold: \textit{P131}\quad Baseline models: \textit{N/A}\quad +\,KIRE: \textit{P131}} \vspace{2mm}\\
		\hline
		$[S1]$ & \multicolumn{3}{p{.95\columnwidth}}{
        The news that British's Prince Harry is engaged to his partner \textit{\textbf{\textcolor{cyan}{Meghan Markle}}} has attracted widespread attention from England, \textit{\textbf{\textcolor{magenta}{America}}} and around the world.} \\
		$[S10]$ & \multicolumn{3}{p{.95\columnwidth}}{\textit{\textbf{\textcolor{cyan}{Markle}}}'s parents \textit{\textbf{\textcolor{magenta}{Thomas Markle}}} and Doria Ragland said in a statement: ...} \\
		\hline
		\multicolumn{4}{p{1\columnwidth}}{\small \textbf{Case 2}\quad Gold: \textit{citizen\_of}\quad Baseline models: \textit{N/A}\quad +\,KIRE: \textit{citizen\_of}} \vspace{2mm}\\
		\hline
		$[S1]$ & \multicolumn{3}{p{.95\columnwidth}}{\textit{\textbf{\textcolor{cyan}{Robert Kingsbury Huntington}}} ... was a naval aircrewman and member of Torpedo Squadron 8 (or VT-8).} \\ 
		$[S2]$ & \multicolumn{3}{p{.95\columnwidth}}{\textit{\textbf{\textcolor{cyan}{He}}} was radioman/gunner to \textit{\textbf{\textcolor{magenta}{Ensign George Gay}}}'s TBD Devastator aircraft ...} \\
		$[S4]$ & \multicolumn{3}{p{.95\columnwidth}}{Born in Los Angeles ... \textit{\textbf{\textcolor{cyan}{he}}} was enlisted in \textit{\textbf{\textcolor{magenta}{the United States Navy}}} 21 Apr. 1941.} \\
		\hline
		\multicolumn{4}{p{1\columnwidth}}{\small \textbf{Case 3}\quad Gold: \textit{P241}\quad Baseline models: \textit{N/A}\quad +\,KIRE: \textit{N/A}} \\	
	\end{tabular}}
\end{table}

\section{Conclusion}
\label{sect:concl}

In this paper, we propose KIRE, an entity knowledge injection framework for enhancing document-level RE.
Coreference knowledge is injected by coreference distillation, while factual knowledge is injected and fused with document representations via representation reconciliation.
Our experiments validate the generalization and the stable performance increase of KIRE to various RE models.
For future work, we plan to exploit other knowledge injection frameworks and integrate more knowledge sources.

\medskip
\noindent\textit{Supplemental Material Statement:} Source code for KIRE is available from Github at \url{https://github.com/nju-websoft/KIRE}.
Datasets are available from \cite{yao2019docred,zaporojets2021dwie}.
%
%
%
\bibliographystyle{splncs04}
\bibliography{iswc2022}

\end{document}